\documentclass[letterpaper]{article} 
\usepackage{aaai2027}  

\usepackage[hyphens]{url}  
\usepackage{graphicx} 
\usepackage{natbib}  
\usepackage{caption} 
\usepackage{algorithm}
\usepackage{amsfonts}

\usepackage{newfloat}
\usepackage{listings}
\DeclareCaptionStyle{ruled}{labelfont=normalfont,labelsep=colon,strut=off} 
\floatstyle{ruled}
\newfloat{listing}{tb}{lst}{}
\floatname{listing}{Listing}

\usepackage{booktabs}

\usepackage[caption=false]{subfig}

\usepackage{multirow}
\usepackage{algpseudocode}
\usepackage{amsmath}

\nocopyright

\title{FRACTION: Fraud Explanation Generation via Agentic Graph \\Context Navigation}
\author{
    Cecelia Soh\textsuperscript{\rm 1}\equalcontrib\thanks{Work done during internship at ByteDance Inc.}, Jun Hu\textsuperscript{\rm 3}\equalcontrib, Yuan Li\textsuperscript{\rm 2}\corresponding, Cheng Chen\textsuperscript{\rm 2}, Hui Liu\textsuperscript{\rm 2}, \\Yanjia Zhu\textsuperscript{\rm 2}, Hao He\textsuperscript{\rm 4}, Bryan Hooi\textsuperscript{\rm 1}, Bingsheng He\textsuperscript{\rm 1}
}
\affiliations{
    \textsuperscript{\rm 1}National University of Singapore\\
    \textsuperscript{\rm 2}ByteDance Inc.\\
    \textsuperscript{\rm 3}Hefei University of Technology\\
    \textsuperscript{\rm 4}Virginia Tech\\

}

\begin{document}

\maketitle

\begin{abstract} 
Graph fraud explanation generation emerges as an important task, which requires jointly leveraging graph structure and content information to produce human-readable explanations for a given risk label.
An intuitive approach is to adapt existing graph-enhanced large language models (LLMs), which inject graph structure into LLMs via graph-conditioned instruction tuning or structure-aware embeddings.
However, existing methods face two key challenges in graph fraud explanation generation:
a lack of evaluation protocols tailored to explanation quality, and 
a mismatch between the granularity of graph information they provide and what explanation generation requires.
In this paper, we propose \textbf{FR}aud Explanation Generation via \textbf{A}gentic Graph \textbf{C}ontext Naviga\textbf{TION} (\textsc{FRACTION}), a framework for graph fraud explanation generation.
To evaluate explanation quality, we introduce a task-specific rubric that assesses faithfulness, graph grounding, helpfulness, consistency, and conciseness, and validate it against human judgments.
To focus on fine-grained details of relevant neighbors without reading the entire neighborhood, we equip a language model with graph tools that progressively disclose neighbor records.
To train our graph agent, we propose a rubric-guided trajectory distillation strategy through which the agent learns both tool-use behavior and structured explanation generation.
Experimental results on three fraud datasets demonstrate consistent improvements over state-of-the-art baselines under a human-aligned rubric.
\end{abstract}

\section{Introduction} \label{sec:introduction}

\begin{figure}[t]
    \centering
    \includegraphics[width=\linewidth]{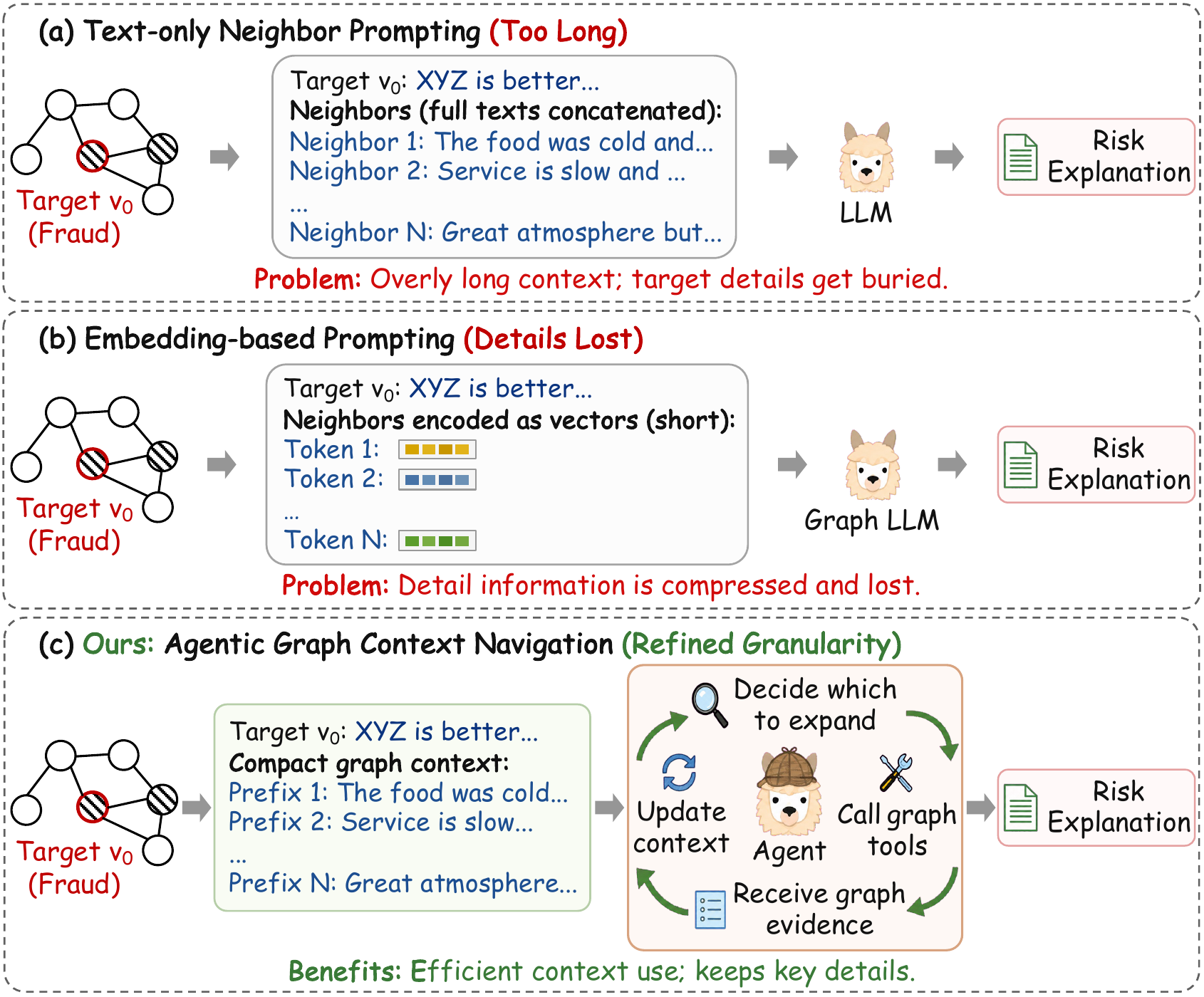}
    \caption{Comparison of graph-to-prompt methods for graph-based fraud risk explanation.}
    \label{fig:teaser}
\end{figure}

Graph-based fraud analysis has become increasingly important, with graph data capturing rich relational information among entities such as users, reviews, and companies~\cite{cheng2025gnnfraudreview, irofti2019graphfraudsurvey}.
Typically, in graph data, nodes represent entities, while edges encode relationships among them, such as users reviewing the same products, accounts sharing common devices, or reviews written by the same author, all of which provide valuable signals for identifying fraudulent behavior.
Studies on tasks such as graph-based fraud detection~\cite{zhuo2024pmp,chen2024consisgad,yu2024bsl,duan2025gaap} have demonstrated that leveraging relational information can significantly improve model performance.

\textbf{Graph Fraud Explanation Generation} is a task that generates human-readable explanations for fraud detection results based on graph data.
It considers not only the target node itself but also relational evidence from its neighboring nodes.
For example, consider a review graph in which nodes represent reviews and edges indicate relations such as reviews written by the same user or posted for the same product. Given a review labeled as fraudulent, the generated explanation may combine evidence from both its content and its neighborhood, e.g., ``the target review contains malicious language, and a neighboring review written by the same author exhibits similar suspicious content.''

However, most existing graph fraud analysis studies focus on tasks such as fraud detection, with limited attention paid to explanation generation.
An intuitive approach is to adapt graph-enhanced LLMs~\cite{chen2024llaga,li2025dgp} that take graph structure and text as joint input and generate human-readable explanations with LLMs.
One line of work represents graph information directly as natural-language input to LLMs, as shown in Figure~\ref{fig:teaser} (a).
For example, TLG~\citep{fatemi2024talk} and GraphText~\cite{zhao2024graphtext} encode graph-structured information as textual prompts for LLMs.
Another line of work instead encodes graph structure into dense representations that are injected into the LLM: for example, LLaGA~\citep{chen2024llaga} and GraphGPT~\citep{tang2024graphgpt} reorganize neighborhoods into structure-aware embedding sequences, as shown in Figure~\ref{fig:teaser} (b).
Despite their effectiveness, these graph-enhanced LLMs are primarily designed and optimized for tasks such as node classification and link prediction, without addressing two key challenges in fraud explanation generation.

The first challenge lies in evaluating open-ended fraud explanations. Unlike node classification or link prediction, whose outputs can be directly compared against fixed ground-truth labels, fraud explanations do not have a unique textual ground truth, as multiple explanations may express the same conclusion in distinct ways. Consequently, lexical-overlap metrics cannot adequately evaluate semantically equivalent explanations with different wording. Moreover, evaluating fraud explanations requires considering multiple task-specific aspects, including the use of target attributes, graph relations, and neighboring records, as well as the usefulness, consistency, and conciseness of the explanation.

The second challenge is that the way existing graph-enhanced LLMs represent graph information may not match the granularity required by fraud explanation generation.
Producing a high-quality explanation for a target node requires focus on fine-grained details from its relevant neighbors, so that each claim in the explanation can be traced back to specific evidence.
However, text-based methods that place neighbor information directly into the prompt either overwhelm this focus when including all neighbors or lose it entirely when relying on a fixed retrieval budget that omits the records that matter most; embedding-based methods compress each neighboring node or subgraph into a fixed-dimensional representation, which discards the textual and metadata details needed to justify a specific conclusion.

In light of these challenges, we propose \textbf{FR}aud Explanation Generation via \textbf{A}gentic Graph \textbf{C}ontext Naviga\textbf{TION} (\textsc{FRACTION}), a framework for graph fraud explanation generation.
To evaluate explanation quality, we introduce a task-specific rubric that assesses faithfulness, graph grounding, helpfulness, consistency, and conciseness, and validate it against human judgments.
To focus on fine-grained details of relevant neighbors without reading the entire neighborhood, we equip a language model with graph tools that progressively disclose neighbor records.
To train our graph agent, we propose a rubric-guided trajectory distillation strategy through which the agent learns both tool-use behavior and structured explanation generation.
Experimental results on three fraud datasets demonstrate consistent improvements over state-of-the-art baselines under a human-aligned rubric.

Our main contributions are as follows:
\begin{itemize}
\item We propose FRACTION, a framework for graph fraud explanation generation, together with a task-specific rubric that assesses faithfulness, graph grounding, helpfulness, consistency, and conciseness, and validate it against human judgments.
\item To focus on fine-grained details of relevant neighbors without reading the entire neighborhood, we equip a language model with graph tools that progressively disclose neighbor records. 
To train this graph agent, we propose a rubric-guided trajectory distillation strategy through which the agent learns both tool-use behavior and structured explanation generation.
\item Experimental results on three fraud datasets demonstrate consistent improvements over state-of-the-art baselines under a human-aligned rubric.
\end{itemize}

\section{Related Work} \label{sec:related-work}

\subsection{Graph-Enhanced Large Language Models}

Recent studies have extended LLMs to graph-structured data by incorporating both graph structure and node attributes into their inputs. Existing methods can be broadly categorized by how graph context is exposed to the LLM. One line of work represents graph information explicitly as textual context. TLG~\citep{fatemi2024talk} studies textual encodings of graph structures, GraphText~\citep{zhao2024graphtext} constructs graph-syntax trees and linearizes them into text sequences, and InstructGLM~\citep{ye2024instructglm} describes multi-scale graph neighborhoods through natural-language instructions. To make such context more manageable, G-Retriever~\citep{he2024gretriever} retrieves a task-relevant subgraph and provides its textualized content, together with a learned graph prompt, to the LLM, while DGP~\citep{li2025dgp} constructs text-only prompts that preserve fine-grained target-node information and summarize neighboring content.

Another line of work maps graph structure into dense representations that are injected into the LLM. GraphGPT~\citep{tang2024graphgpt} aligns graph representations with text through graph instruction tuning, while LLaGA~\citep{chen2024llaga} reorganizes graph neighborhoods into structure-aware embedding sequences and projects them into the LLM input space. Despite their effectiveness, these methods expose graph context that is determined before generation and are primarily developed for graph prediction or question answering. Neither paradigm directly supports the progressive acquisition of verifiable neighbor evidence required for fraud explanation generation.

\subsection{Natural-Language Explanations for Graph-Based Fraud Analysis}

Existing explainable graph-based models for fraud detection and related
applications expose influential features, relations, neighboring nodes, or
message-passing structures through attention, expert knowledge, post-hoc
explanation, architecture search, or learned masks. Early attention-based
methods such as SemiGNN~\citep{wang2019semignn} and AMG-DP~\citep{hu2020loan} identify important neighbors, graph views,
and relation types. Know-GNN~\citep{rao2021knowgnn} represents
expert knowledge as graph functional dependencies to guide detection and
provide rule-based explanations, whereas NGS~\citep{qin2022ngs} searches for interpretable
meta-graphs that expose the relational message-passing structures used for
prediction. xFraud~\citep{rao2021xfraud} combines a heterogeneous
graph detector with a post-hoc explainer, MultiFraud~\citep{wu2024multifraud} produces explanations
across multiple heterogeneous graphs, and SEFraud~\citep{li2024sefraud} jointly learns feature and
edge masks for self-explainable fraud detection. Although these methods
reveal which graph components, relational patterns, or domain rules contribute
to a prediction, they do not directly synthesize fine-grained textual evidence
into an auditor-facing natural-language report.

Recent work has explored natural-language explanations for model predictions. GraphNarrator~\citep{pan2025graphnarrator} constructs saliency-derived pseudo-explanations and trains an end-to-end generator to explain GNN predictions. Beyond graph models, PEX~\citep{zhao2025pex} measures prediction--explanation consistency and uses preference optimization to improve it; SR-NLE~\citep{wang2025srnle} improves post-hoc explanations through iterative self-critique and refinement; and FaithLM~\citep{chuang2026faithlm} evaluates and optimizes faithfulness through intervention-based refinement. Different from existing work, our FRACTION framework addresses graph-grounded fraud explanation through agentic graph context navigation, allowing the model to progressively acquire fine-grained evidence from a small set of relevant neighbors and ground each claim in verifiable records.

\begin{figure*}[t]
    \centering
    \includegraphics[width=0.9\linewidth]{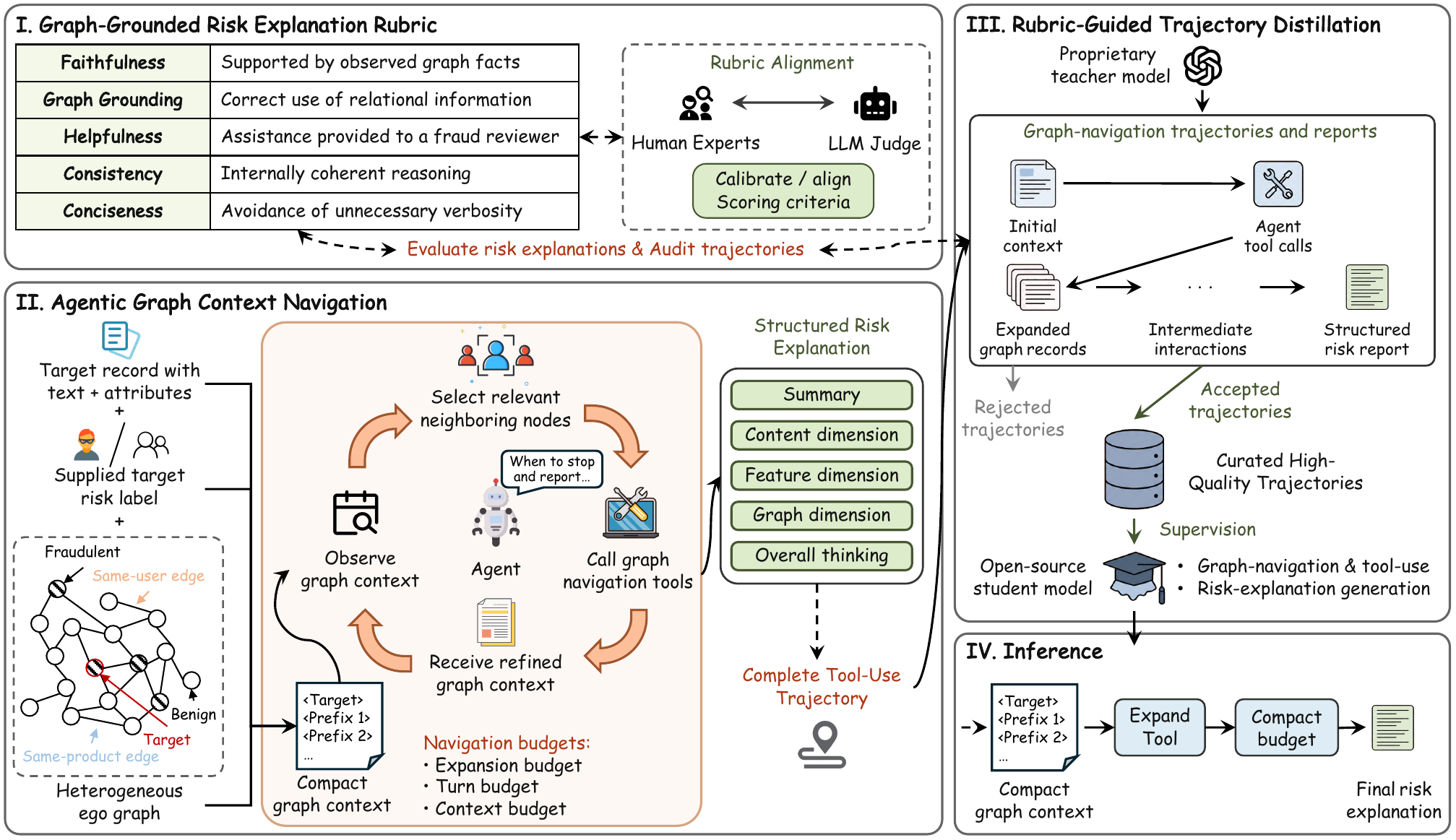}
    \caption{Overall framework of \textbf{FR}aud Explanation Generation via \textbf{A}gentic Graph \textbf{C}ontext Naviga\textbf{TION} (\textsc{FRACTION}).}
    \label{fig:framework}
\end{figure*}

\section{Problem Definition}

Let $\mathcal{G}=(\mathcal{V},\mathcal{E},\mathcal{R})$ be a heterogeneous graph. A target node $v\in\mathcal{V}$ has textual and scalar attributes $x_v$ (e.g., review text and rating), a supplied label $y_v\in\{\textsc{Yes},\textsc{No}\}$, and relation-specific neighbors $\mathcal{N}_r(v)$ for $r\in\mathcal{R}$. The system generates a human-readable report:
\begin{equation}
z_v=f_\theta\bigl(x_v,y_v,\mathcal{C}_0(v),\mathcal{T}_v\bigr),
\end{equation}
where $\mathcal{C}_0(v)$ is the initial graph context and $\mathcal{T}_v$ is an optional sequence of tool interactions. The label is an input to explain instead of the prediction target. 
Following the industrial practice of ByteDance, we require each report to contain the following sections: (i) \textit{Summary} states the main conclusion; (ii) \textit{Content Dimension} analyzes the textual content; (iii) \textit{Feature Dimension} examines the scalar and metadata features; (iv) \textit{Graph Dimension} evaluates evidence from relation-specific neighbors; and (v) \textit{Overall Thinking} integrates the preceding evidence into the final assessment. Together, these sections provide a structured and evidence-based analysis of the target node.

\section{Method} 
\label{sec:method}

In this section, we introduce \textbf{FR}aud Explanation Generation via \textbf{A}gentic Graph \textbf{C}ontext Naviga\textbf{TION} (\textsc{FRACTION}), a novel framework for graph fraud explanation generation.

\subsection{Overall Framework}

As shown in Figure~\ref{fig:framework}, FRACTION consists of the following three main components:
\begin{itemize}
\item \textbf{Graph-Grounded Risk Explanation Rubric}: We introduce a task-specific rubric that assesses faithfulness, graph grounding, helpfulness, consistency, and conciseness, and validate it against human judgments.

\item \textbf{Agentic Graph Context Navigation}: We equip a language-model agent with graph tools that progressively disclose selected neighbor records, enabling access to fine-grained evidence without loading every neighbor record in full.

\item \textbf{Rubric-Guided Trajectory Distillation}: We filter teacher trajectories using the rubric and distill the accepted trajectories into an open-weight student model, which learns both graph-tool use and structured explanation generation.

\end{itemize}

\subsection{Graph-Grounded Risk Explanation Rubric}
\label{sec:rubric}

Existing lexical-overlap metrics against a single reference are inadequate for this task, since multiple valid reports may describe the same evidence using substantially different wording. To evaluate explanation quality, we define a reference-free, evidence-conditioned rubric with five complementary dimensions: 
(a) \emph{Faithfulness}: factual claims are supported by observed evidence; 
(b) \emph{Graph grounding}: neighbors and relations are used correctly;
(c) \emph{Helpfulness}: the report is useful to a human reviewer;
(d) \emph{Consistency}: the report agrees with the label;
(e) \emph{Conciseness}: the report avoids redundancy and irrelevant content.

For an explanation trajectory $\tau$, the judge assigns an integer score $s_m\in\{1,\ldots,5\}$ to each dimension $m$. We aggregate the five scores as
\begin{equation}\label{eq:rubrics}
\begin{array}{rcl}
q(\tau)&=&0.35s_{\mathrm{faith}}+0.25s_{\mathrm{graph}}
          +0.25s_{\mathrm{help}}\\
       &&+0.10s_{\mathrm{consis}}+0.05s_{\mathrm{concise}},
\end{array}
\end{equation}
where the coefficients are provided by ByteDance based on the preferences of domain experts. The judge receives the target record, the full graph context, and the final report, and evaluates each claim against available evidence. We validate the alignment between rubric-based judgments and human assessments, and apply the same rubric criteria to trajectory filtering and final evaluation.
For example, suppose a report claims that the target review is suspicious because several reviews connected through a \emph{Same User} relation use similar language. The claim receives high faithfulness and graph-grounding scores only if the observed records support the stated language pattern and the corresponding relation exists in the graph. A fluent claim that relies on undisclosed record details or an incorrect relation type is penalized, even if its conclusion matches the supplied risk label.
\subsection{Agentic Graph Context Navigation}

To retrieve fine-grained evidence without loading all neighboring records in full, we design \textbf{Agentic Graph Context Navigation (AGCN)}, which equips a language-model agent with graph tools that progressively disclose selected neighbor texts on demand (Figure~\ref{fig:framework}, II).

\paragraph{Compact graph context.}
Given a target node $v$, let $\mathcal{N}_k(v)$ denote the nodes within $k$ hops of $v$, excluding $v$ itself; $g_{v,u}$ the textualized hop and relation information connecting node $v$ to node $u$; and $\operatorname{pref}(x_u)$ a short prefix of its record. After arranging neighbors in a fixed relation-aware order, the initial context is

\begin{equation}
\label{eq:initial-context}
\mathcal{C}_0(v)
=
x_v
\oplus
y_v
\oplus
\bigoplus_{u\in\mathcal{N}_k(v)}
\bigl\langle
u,\,g_{v,u},\,\operatorname{pref}(x_u)
\bigr\rangle ,
\end{equation}
where $\oplus$ denotes concatenation.
This representation exposes the breadth and heterogeneous structure of the $k$-hop ego graph while deferring the retrieval of more detailed node texts. AGCN therefore selects evidence from an informed, preconstructed neighborhood.

\paragraph{Graph navigation tools.}
At turn $t$, the agent is allowed to select visible node IDs $S_t\subseteq\mathcal{N}_k(v)$ and invoke one of three tools:
\textsc{FullText}$(S_t)$ returns complete texts;
\textsc{TruncatedText}$(S_t,\ell_t)$ returns texts truncated to length
$\ell_t$; and 
\textsc{TextLength}$(S_t)$ returns text lengths to guide the choice of retrieval granularity. 
Together, these tools allow the agent to adapt the amount of disclosed evidence under a limited context budget.

\paragraph{Graph navigation loop.}
At each turn, the agent either invokes a tool or stops to generate the final report:
\begin{equation}
\begin{aligned}
a_t
&\sim
\pi_\theta(\cdot\mid\mathcal{C}_{t-1},B_t),
\qquad
a_t\in\{a_{\mathrm{stop}}\}\cup\mathcal{A}_{\mathrm{tool}},\\
\mathcal{C}_t
&=
\mathcal{C}_{t-1}\oplus(a_t,o_t),
\qquad
\text{if }a_t\in\mathcal{A}_{\mathrm{tool}},
\end{aligned}
\end{equation}
where $\mathcal{A}_{\mathrm{tool}}$ comprises parameterized calls to \textsc{FullText}, \textsc{TruncatedText}, and \textsc{TextLength}, including the selected node set $S_t$ and, when applicable, the truncation length $\ell_t$. 
$B_t$ denotes the remaining navigation budgets, $\oplus$ denotes context concatenation, and $o_t$ denotes the tool output. The process terminates when the agent considers the accumulated evidence sufficient or when no further tool call is permitted by the budgets. The agent then generates a structured risk explanation from the resulting context. The tool calls, tool outputs, and final explanation are retained as a trajectory for later distillation learning.

\subsection{Rubric-Guided Trajectory Distillation}

To train the graph agent, we distill rubric-filtered trajectories generated by a large teacher model into a smaller open-weight student model. For each training target, the teacher generates a trajectory:
\begin{equation}
\tau=
\left(
\mathcal{C}_0,
a_1,o_1,\ldots,
a_{n_\tau},o_{n_\tau},
z
\right),
\end{equation}
where $n_\tau$ is the number of tool calls, $a_t$ and $o_t$ are the corresponding tool action and output, and $z$ is the final explanation.

We evaluate each teacher trajectory using a separate LLM judge under the graph-grounded rubric. 
Let $\mathcal{D}_{\mathrm{cand}}$ denote the set of candidate trajectories generated by the teacher. 
We first retain trajectories that meet the rubric quality threshold:
\begin{equation}
\mathcal{D}_{q}
=
\left\{
\tau\in\mathcal{D}_{\mathrm{cand}}
\mid q(\tau)\geq\gamma
\right\},
\end{equation}
where $\mathcal{D}_{q}$ is the quality-filtered trajectory set, $q(\tau)$ is defined in Equation~\ref{eq:rubrics}, and $\gamma$ is the quality threshold. We then apply structural constraints to construct the final distillation dataset:
\begin{equation}
\mathcal{D}_{\mathrm{keep}}
=
\left\{
\tau\in\mathcal{D}_{q}
\;\middle|\;
\begin{aligned}
n_{\mathrm{exp}}(\tau)&\in[n_{\min},n_{\max}],\\
\ell_{\mathrm{rep}}(\tau)&\in[\ell_{\min},\ell_{\max}],\\
c_{\mathrm{sec}}(\tau)&=1
\end{aligned}
\right\},
\end{equation}
where $n_{\mathrm{exp}}(\tau)$ denotes the number of neighbor-expansion rounds in trajectory $\tau$, $\ell_{\mathrm{rep}}(\tau)$ denotes the final report length, and $c_{\mathrm{sec}}(\tau)$ indicates whether all required report sections are present and correctly ordered. 
The resulting $\mathcal{D}_{\mathrm{keep}}$ is the final set of teacher trajectories retained for student fine-tuning after quality and structural filtering.

Rather than supervising only the final explanation, we distill the complete accepted trajectory. For a tokenized trajectory $(w_1,\ldots,w_{T_\tau})$, let $\mathcal{I}_{\mathrm{asst}}(\tau)$ denote the positions generated by the assistant, including graph calls and the final explanation.
The supervision mask is
\begin{equation}
m_i^\tau
=
\mathbb{I}
\left[
i\in\mathcal{I}_{\mathrm{asst}}(\tau)
\right],
\end{equation}
and the student is optimized with
\begin{equation}
\mathcal{L}_{\mathrm{SFT}}
=
-\sum_{\tau\in\mathcal{D}_{\mathrm{keep}}}
\sum_{i=1}^{T_\tau}
m_i^\tau
\log p_\theta(w_i\mid w_{<i}).
\end{equation}
The initial graph context and tool outputs are masked from the loss because they are provided by the environment. The student conditions on these spans while being supervised to generate both graph calls and the final structured explanation.

\subsection{Inference}

At deployment, the student receives the same initial context, tool schema, and navigation budgets as the teacher. It independently selects neighbors, conditions on the tool outputs returned during navigation, and generates the final structured explanation without further calls to the teacher.

\section{Experiments}

We evaluate FRACTION on three graph-based fraud datasets.
DeepSeek-v4-Pro~\cite{xu2026deepseek} is used for teacher-trajectory generation, while GPT-5.5~\cite{singh2025openai} serves as the LLM judge. All methods are provided with the target risk label, so the experiments evaluate explanation generation rather than fraud classification.

\subsection{Datasets}
We conduct experiments on YelpReviews, AmazonVideo, and AdGov, each of which contains text-rich nodes, risk labels, and application-specific graph relations. The detailed statistics are shown in Table~\ref{tab:dataset-statistics}. For each dataset, we use 1,000 training targets, 200 validation targets, and 400 test targets.

\textbf{YelpReviews} is derived from YelpChi~\citep{rayana2015collective}, where reviews are represented as nodes and edges encode \emph{Same User},
\emph{Same Business and Rating}, and \emph{Same Business and Month} relations.
\textbf{AmazonVideo} is constructed from the Amazon review corpus
\citep{mcauley2015inferring}. Its edges represent \emph{Same Product},
\emph{Same User}, and \emph{Same Week and Rating} relations. A review is
labeled risky when it receives at least one helpfulness vote and its ratio of
helpful votes to total votes is below 0.5; all other reviews are labeled
non-risky.
\textbf{AdGov} is an industrial graph collected from a proprietary advertising risk-governance platform. Its nodes represent daily company snapshots containing company attributes and account activity. Edges connect companies through shared or similar identity information, such as business-registration attributes.

\begin{table}[t]
    \centering
    \scalebox{0.75}{
    \begin{tabular}{llrrc}
        \toprule
        \multirow{2}{*}{\textbf{Dataset}}
        & \multirow{2}{*}{\textbf{Node Type}}
        & \multirow{2}{*}{\textbf{\# Nodes}}
        & \multirow{2}{*}{\textbf{\# Edges}}
        & \textbf{\# Edge} \\
        & & & & \textbf{Types} \\
    \midrule
    YelpReviews & Service Review    & 67,395 & 17,486,608 & 3 \\
    AmazonVideo & Product Review    & 37,126 & 9,883,406  & 3 \\
    AdGov       & Company Snapshot  & 2,126,463  & 24,240,638     & 39 \\
    \bottomrule
\end{tabular}%
    }
\caption{Statistics of the three datasets.}
\label{tab:dataset-statistics}
\end{table}

\begin{table*}[t]
\centering

\renewcommand{\arraystretch}{1.15}
\setlength{\tabcolsep}{5pt}

\scalebox{0.75}{
\begin{tabular}{llccccccc}
\toprule

Method
& Model
& Final score
& Pass rate
& Faithfulness
& Graph grounding
& Helpfulness
& Consistency
& Conciseness \\
\midrule

\multicolumn{9}{c}{\textit{Dataset: YelpReviews}} \\
\midrule

Zero-shot target-only
& Qwen3.6-27B
& 2.0615 & 0.00\% & 1.9300 & 1.0000 & 2.0425 & 3.9425 & 4.6225 \\

Zero-shot target-only
& DeepSeek-v4-Pro
& 2.2420 & 0.00\% & 2.1200 & 1.0000 & 2.3450 & 4.2375 & 4.8000 \\

TLG
& Qwen3.6-27B
& 1.9266 & 0.00\% & 1.7075 & 1.0025 & 1.8975 & 3.7025 & 4.6750 \\

TLG
& DeepSeek-v4-Pro
& 1.9545 & 0.00\% & 1.7500 & 1.0025 & 1.9225 & 3.7825 & 4.6500 \\

GraphText
& Qwen3.6-27B
& 2.0855 & 1.75\% & 1.8800 & 1.1625 & 2.0800 & 3.8500 & 4.6375 \\

GraphText
& DeepSeek-v4-Pro
& 2.1558 & 2.00\% & 1.9450 & 1.1425 & 2.1750 & 4.0850 & 4.7425 \\

InstructGLM
& Qwen3.6-27B
& 2.4254 & 12.75\% & 2.0950 & 1.6050 & 2.5425 & 4.2825 & 4.5400 \\

DGP
& Qwen3.6-27B
& 2.6930 & 37.00\% & 2.4400 & 1.8175 & 2.8225 & 4.4575 & 4.6650 \\

LLaGA
& Qwen3.6-27B
& 2.4921 & 17.50\% & 2.1400 & 1.7450 & 2.6200 & 4.2525 & 4.5325 \\

PEX
& Qwen3.6-27B
& 2.2126 & 0.00\% & 2.0450 & 1.0000 & 2.3350 & 4.2525 & 4.7575 \\

GraphNarrator
& Qwen3.6-27B
& 2.1933 & 3.50\% & 1.8675 & 1.4875 & 2.2600 & 4.0050 & 4.0450 \\

\midrule

FRACTION
& Qwen3.6-27B
& \textbf{3.7080} & \textbf{91.75\%} & \textbf{3.2450}
& \textbf{3.3625} & \textbf{3.9925} & \textbf{4.9200}
& \textbf{4.8300} \\

\midrule

\multicolumn{9}{c}{\textit{Dataset: AmazonVideo}} \\
\midrule

Zero-shot target-only
& Qwen3.6-27B
& 2.0826 & 0.00\% & 1.9525 & 1.0000 & 2.0600 & 4.0375 & 4.6100 \\

Zero-shot target-only
& DeepSeek-v4-Pro
& 2.1450 & 0.00\% & 2.0125 & 1.0000 & 2.1150 & 4.2450 & 4.7475 \\

TLG
& Qwen3.6-27B
& 1.9259 & 0.00\% & 1.6925 & 1.0150 & 1.9050 & 3.7700 & 4.5300 \\

TLG
& DeepSeek-v4-Pro
& 1.9121 & 0.25\% & 1.6375 & 1.0075 & 1.8850 & 3.8575 & 4.6025 \\

GraphText
& Qwen3.6-27B
& 2.1226 & 3.75\% & 1.9400 & 1.1525 & 2.0850 & 4.0125 & 4.6600 \\

GraphText
& DeepSeek-v4-Pro
& 2.1090 & 1.50\% & 1.9400 & 1.0900 & 2.0775 & 4.0325 & 4.6975 \\

InstructGLM
& Qwen3.6-27B
& 2.4825 & 18.75\% & 2.2125 & 1.6375 & 2.5700 & 4.2950 & 4.5350 \\

DGP
& Qwen3.6-27B
& 2.5631 & 23.75\% & 2.3050 & 1.6650 & 2.7000 & 4.4025 & 4.4975 \\

LLaGA
& Qwen3.6-27B
& 2.5264 & 21.75\% & 2.2000 & 1.7875 & 2.5725 & 4.4200 & 4.4875 \\

PEX
& Qwen3.6-27B
& 2.2261 & 0.00\% & 2.0750 & 1.0000 & 2.2550 & 4.5025 & 4.7175 \\

GraphNarrator
& Qwen3.6-27B
& 2.0783 & 1.50\% & 1.8275 & 1.2075 & 2.1025 & 4.1200 & 3.9825 \\

\midrule

FRACTION
& Qwen3.6-27B
& \textbf{4.1745} & \textbf{98.75\%} & \textbf{4.2225}
& \textbf{3.5675} & \textbf{4.2500} & \textbf{4.9575}
& \textbf{4.9300} \\

\midrule

\multicolumn{9}{c}{\textit{Dataset: AdGov}} \\
\midrule

Zero-shot target-only
& DeepSeek-v4-Pro
& 2.7900 & 35.25\% & 2.5575 & 1.8425 & 3.0650 & 4.2900 & 4.7800 \\

Zero-shot target-only
& Qwen3.6-27B
& 2.7988 & 36.25\% & 2.5050 & 1.9075 & 3.1050 & 4.3250 & 4.7275 \\

TLG
& Qwen3.6-27B
& 2.8809 & 32.00\% & 2.3350 & 2.3450 & 3.2075 & 4.4400 & 4.6300 \\

TLG
& DeepSeek-v4-Pro
& 2.8565 & 29.00\% & 2.2950 & 2.3575 & 3.2275 & 4.2250 & 4.6900 \\

GraphText
& Qwen3.6-27B
& 3.0145 & 42.00\% & 2.4500 & 2.5525 & 3.3325 & 4.5125 & 4.6900 \\

GraphText
& DeepSeek-v4-Pro
& 2.8566 & 30.75\% & 2.3125 & 2.4500 & 3.1900 & 4.0225 & 4.7000 \\

InstructGLM
& Qwen3.6-27B
& 2.7996 & 27.50\% & 2.2475 & 2.2875 & 3.0875 & 4.3500 & 4.6850 \\

DGP
& Qwen3.6-27B
& 2.8025 & 31.50\% & 2.3575 & 2.1600 & 3.0675 & 4.3450 & 4.7200 \\

LLaGA
& Qwen3.6-27B
& 3.0299 & 46.00\% & 2.4475 & 2.6425 & 3.3150 & 4.5500 & 4.5775 \\

PEX
& Qwen3.6-27B
& 2.4074 & 0.25\% & 2.3425 & 1.0050 & 2.7350 & 4.1475 & 4.7550 \\

GraphNarrator
& Qwen3.6-27B
& 2.1253 & 7.75\% & 1.9200 & 1.2525 & 2.4125 & 3.1900 & 4.3600 \\

\midrule

FRACTION
& Qwen3.6-27B
& \textbf{4.1035} & \textbf{94.25\%} & \textbf{4.0125}
& \textbf{3.6300} & \textbf{4.2250} & \textbf{4.9075}
& \textbf{4.8925} \\

\bottomrule
\end{tabular}%
}

\caption{Performance comparison on graph fraud explanation generation across YelpReviews, AmazonVideo, and AdGov.}
\label{tab:baseline}
\end{table*}

\subsection{Baselines}

We compare FRACTION with four groups of baselines: target-only prompting with
Qwen3.6-27B~\cite{yang2025qwen3} and DeepSeek-v4-Pro~\cite{xu2026deepseek}; static graph-to-text methods, including TLG
and GraphText; graph-enhanced LLMs, including LLaGA
\citep{chen2024llaga}, InstructGLM~\citep{ye2024instructglm}, and DGP
\citep{li2025dgp}; and explanation-generation methods, including
GraphNarrator~\citep{pan2025graphnarrator} and PEX~\citep{zhao2025pex}.
We use official implementations and recommended
configurations whenever available.

\subsection{Implementation Details}

Each initially visible neighbor is represented by a 220-character prefix, with at most 100 neighbors per relation. The graph navigation budget permits up to 100 expanded nodes over 10 interaction turns. 
For each training target, we sample up to 10 teacher trajectories using a temperature of 0.7 and top-$p$ of 0.9, stopping once one is admitted to $\mathcal{D}_{\mathrm{keep}}$. We set $\gamma=4.0$, $k=1$, $[n_{\min},n_{\max}]=[0,10]$, and $[\ell_{\min},\ell_{\max}]=[350,5{,}000]$ characters.

The student is initialized from Qwen3.6-27B and fine-tuned for two epochs using 4-bit LoRA with rank 16, scaling factor 32, and dropout 0.05. We use paged 8-bit AdamW with a learning rate of $2\times10^{-4}$ and select the best checkpoint based on validation loss. 

Training and inference use four high-performance GPUs with 96\,GB of memory each. We use seed 42 for data partitioning, target sampling, neighbor ordering, and other local stochastic operations. All methods use the same test targets, prompts, and decoding settings.

\subsection{Evaluation Protocol}

We evaluate generated explanations using the graph-grounded rubric introduced in the Graph-Grounded Risk Explanation Rubric subsection. 
We report Faithfulness, Graph Grounding, Helpfulness, Consistency, and Conciseness, together with their weighted final score and the pass rate under a threshold of 3.0 out of 5.0.

To ensure a fair comparison, all methods are evaluated under the same unified-evidence protocol. The GPT-5.5 judge receives the complete target-node information and all accessible neighbors. Thus, evaluation is based on the same observable evidence regardless of the context or navigation strategy used by each method.

We additionally compare end-to-end inference latency on YelpReviews and AmazonVideo to assess the computational efficiency of different methods.

\subsection{Performance Analysis} Table~\ref{tab:baseline} compares our method with the baselines on YelpReviews, AmazonVideo, and AdGov. We draw the following observations:
\begin{itemize}
\item Zero-shot methods include Zero-shot target-only, TLG, and GraphText. The first uses only the target node, while the latter two are graph-enhanced LLM methods that additionally incorporate graph information. However, their final scores alternately outperform one another across datasets, indicating that zero-shot graph-enhanced methods cannot reliably exploit graph information for this task.

\item Trained graph-enhanced LLMs, including InstructGLM, DGP, and LLaGA, outperform the zero-shot methods in most cases, indicating the importance of training for effectively utilizing graph information in this task.

\item Explanation-generation methods, including PEX and GraphNarrator, consistently underperform the trained graph-enhanced LLMs. PEX relies only on target-node information, while GraphNarrator selects neighbors using predefined saliency signals. One potential reason for GraphNarrator's weaker performance is that these signals do not reliably capture semantic relevance, making it difficult to focus on the neighbor information most useful for explanation generation. This shows that these explanation-generation methods are not directly suited to graph fraud explanation generation.
\item Different from existing methods, FRACTION uses agentic graph context navigation to progressively disclose relevant neighbor records and focus on their fine-grained details. It significantly outperforms all baselines across the three datasets and achieves the highest final scores, demonstrating the effectiveness of agentic graph context navigation for graph fraud explanation generation.

\end{itemize}

\begin{figure}[t]
    \centering
    \includegraphics[width=0.9\columnwidth]{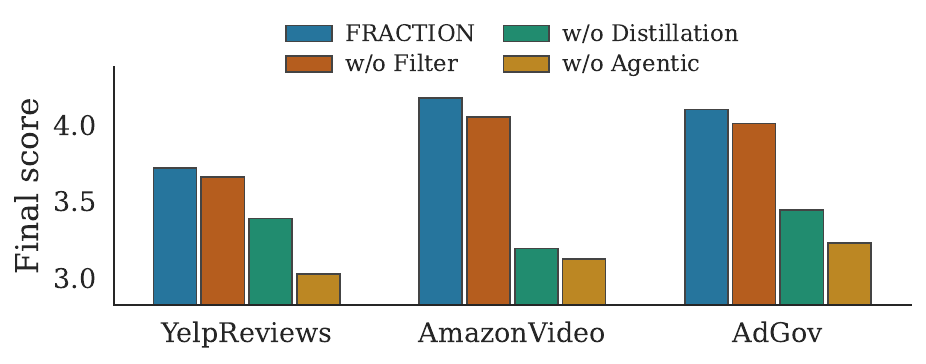}
    \caption{Ablation results of FRACTION and its variants.}
    \label{fig:ablation-study}
\end{figure}

\begin{figure}[t]
    \centering
    \includegraphics[width=0.9\columnwidth]{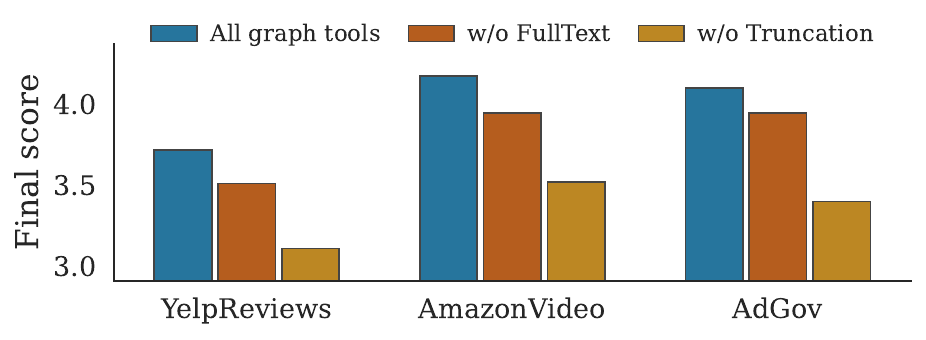}
    \caption{Impact of graph tools.}
    \label{fig:graph-tool-removal}
\end{figure}

\subsection{Detailed Analysis} We further examine the contribution of the main components, the behavior of graph-tool navigation, and the performance--efficiency trade-off.

\subsubsection{Ablation Study}
We compare FRACTION with three variants: \textbf{w/o Agentic}, which replaces
iterative graph navigation with single-turn generation from a fixed graph
context; \textbf{w/o Distillation}, which directly uses the pretrained
Qwen3.6-27B model as a zero-shot graph agent; and \textbf{w/o Filtering},
which retains all teacher trajectories during fine-tuning without quality
filtering. All variants use the same backbone and evaluation settings.

As shown in Figure~\ref{fig:ablation-study}, removing agentic navigation causes
the largest performance drop on all datasets, demonstrating the importance of
progressive evidence acquisition. Removing distillation also substantially
degrades performance, particularly on AmazonVideo, showing that trajectory
supervision is critical for transferring the teacher's navigation behavior.
Filtering provides a smaller but consistent improvement, confirming the value
of removing low-quality trajectories before fine-tuning.

\subsubsection{Impact of Graph Tools}
We evaluate the contribution of different graph navigation tools while keeping the model, prompt, and context budget fixed. Figure~\ref{fig:graph-tool-removal} compares FRACTION with all tools against \textbf{w/o FullText}, which removes the \textsc{FullText} tool, and \textbf{w/o Truncation}, which removes both \textsc{TruncatedText} and \textsc{TextLength}. We remove them together because they are introduced specifically to guide truncated retrieval.

Removing \textbf{FullText} consistently reduces the final score across all datasets, indicating that complete neighbor records provide complementary fine-grained evidence. Removing \textbf{Truncation} causes a substantially larger drop, demonstrating the importance of adaptively controlling evidence granularity under a limited context budget. Overall, truncated retrieval is the primary contributor, while full-text retrieval provides additional detail when needed.

\subsubsection{Efficiency Analysis}
\begin{figure}[t]
    \centering
    \subfloat[YelpReviews\label{fig:efficiency-tradeoff-yelp}]{%
        \includegraphics[width=0.48\columnwidth]
        {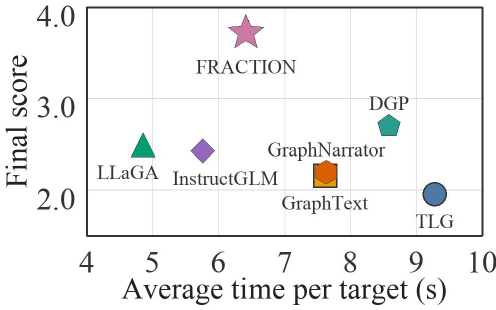}
    }
    \hfill
    \subfloat[AmazonVideo\label{fig:efficiency-tradeoff-amazon}]{%
        \includegraphics[width=0.48\columnwidth]
        {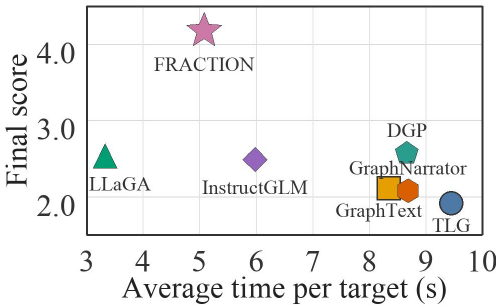}
    }

    \caption{Performance--time trade-off of FRACTION and representative baselines.}
    \label{fig:efficiency-tradeoff}
\end{figure}

Figure~\ref{fig:efficiency-tradeoff} compares explanation quality and average
inference time per target. FRACTION achieves the highest final score on both YelpReviews
and AmazonVideo, requiring 6.42 and 5.08 seconds per target, respectively. It lies
on the empirical Pareto frontier on both datasets, demonstrating that selective
graph navigation substantially improves explanation quality with moderate
inference overhead.

\subsection{Alignment with Human Judgments}

Table~\ref{tab:gpt55-human-alignment} compares GPT-5.5 rubric scores with
human judgments on 20 reports per dataset. GPT-5.5 shows significant score
correlations on YelpReviews ($\rho=0.7205$) and AmazonVideo ($\rho=0.4558$), with
pass-decision agreement of 95\% and 100\%, respectively. These results support
its use as the LLM judge for trajectory filtering and final evaluation.

\begin{table}[t]
\centering
\renewcommand{\arraystretch}{1.1}
\setlength{\tabcolsep}{5pt}
\scalebox{0.68}{
\begin{tabular}{lccccccc}
\toprule
Dataset & $N$ & Spearman $\rho$ & $p$-value & Mean diff. & MAE & RMSE & Pass agree. \\
\midrule
YelpReviews & 20 & 0.7205 & 0.0003 & 0.0500 & 0.2500 & 0.3256 & 95.00\% \\
AmazonVideo & 20 & 0.4558 & 0.0434 & 0.2475 & 0.3625 & 0.4462 & 100.00\% \\

\bottomrule
\end{tabular}%
}
\caption{Alignment of rubric scores with human judgments.}
\label{tab:gpt55-human-alignment}
\end{table}

\section{Conclusion} We study graph fraud explanation generation, which requires jointly leveraging graph structure and content information to produce human-readable explanations for a supplied risk label. We propose FRACTION, a framework for graph fraud explanation generation. To evaluate explanation quality, FRACTION introduces a task-specific rubric that assesses faithfulness, graph grounding, helpfulness, consistency, and conciseness, and validates it against human judgments. To focus on fine-grained details of relevant neighbors without reading the entire neighborhood, FRACTION equips a language model with graph tools that progressively disclose neighbor records. To train the graph agent, FRACTION further adopts a rubric-guided trajectory distillation strategy through which the agent learns both tool-use behavior and structured explanation generation. Experimental results on three fraud datasets demonstrate consistent improvements over state-of-the-art baselines under a human-aligned rubric.

\bibliography{aaai2027}

\end{document}